\documentclass[letterpaper, 10 pt, conference]{ieeeconf}  

\IEEEoverridecommandlockouts                              

\usepackage{graphics} 
\usepackage{epsfig} 
\usepackage{mathptmx} 
\usepackage{times} 
\usepackage{amsmath} 
\usepackage{amssymb}  
\usepackage{color}
\usepackage{xcolor}
\usepackage{caption}

\title{\LARGE \bf
Reinforcement Learning Meets Hybrid Zero Dynamics: \protect\\ A Case Study for RABBIT}
\author{Guillermo A. Castillo$^{1}$, Bowen Weng$^{1}$, Ayonga Hereid$^{2}$ and Wei Zhang$^{1}$
\thanks{*This work of Guillermo A. Castillo, Bowen Weng and Wei Zhang was supported in part by the National Science Foundation under grant CNS-1552838. The work of Ayonga Hereid was supported in part by the Toyota Research Institute (TRI) under award number No.~02281.}
\thanks{$^{1}$Guillermo A. Castillo, Bowen Weng and Wei Zhang are with the Department of Electrical and Computer Engineering, Mathematics and Computer Science, Ohio State University, Columbus, OH 43210, USA
        {\tt\small castillomartinez.2@osu.edu, weng.172@osu.edu, zhang.491@osu.edu}}%
\thanks{$^{2}$Ayonga Hereid is with the Department of Electrical and Computer Engineering, University of Michigan, Ann Arbor, MI 48109, USA  {\tt\small ayonga@umich.edu}}%
}

\graphicspath{{./images/}}

\usepackage[capitalise]{cleveref}

\usepackage[noadjust]{cite}

\usepackage{graphicx}

\usepackage[caption=false]{subfig}

\begin{document}

\maketitle
\thispagestyle{empty}
\pagestyle{empty}

\begin{abstract}
The design of feedback controllers for bipedal robots is challenging due to the hybrid nature of its dynamics and the complexity imposed by high-dimensional bipedal models. In this paper, we present a novel approach for the design of feedback controllers using Reinforcement Learning (RL) and Hybrid Zero Dynamics (HZD). Existing RL approaches for bipedal walking are inefficient as they do not consider the underlying physics, often requires substantial training, and the resulting controller may not be applicable to real robots. HZD is a powerful tool for bipedal control with local stability guarantees of the walking limit cycles. In this paper, we propose a non traditional RL structure that embeds the HZD framework into the policy learning. More specifically, we propose to use RL to find a control policy that maps from the robot's reduced order states to a set of parameters that define the desired trajectories for the robot's joints through the virtual constraints. Then, these trajectories are tracked using an adaptive PD controller. The method results in a stable and robust control policy that is able to track variable speed within a continuous interval. Robustness of the policy is evaluated by applying external forces to the torso of the robot. The proposed RL framework is implemented and demonstrated in OpenAI Gym with the MuJoCo physics engine based on the well-known RABBIT robot model.
\end{abstract}

\section{INTRODUCTION}

Stable dynamic locomotion for bipedal robots is an important problem that has received considerable research attention from the robotics community. It is particularly challenging due to the complexity of high dimensional models, underactuation, unilateral ground contacts, nonlinear and hybrid dynamics, among others. Most existing bipedal walking control methods rely on accurate physical models of the system. These model-based approaches can be further divided into two categories: methods that are based on simplified models and methods that are based on the full order model of the robot.

The linear inverted pendulum (LIP) \cite{Kajita1992} is a popular reduced order model. Since its inception, LIP has been widely used jointly with the zero moment point (ZMP) criteria \cite{Vukobratovic2004} to compute feasible motion trajectories using pattern generators \cite{Yoshida2008}, \cite{Stephens2010}. LIP has also been used along with the Capture Point (CP) approach to analyze the push recovery problem in bipedal walking robots \cite{Pratt2006}, \cite{Pratt2012}. Although the simplicity of reduced models presents many advantages particularly in the practical implementation of the online algorithms, they do not consider the physical joint and actuator limits while designing the gaits, and they often require the robot to be fully actuated.

Another branch of methods uses the full order model of the robot, which can capture the underlying dynamics more accurately, and often leads to more natural dynamic walking behaviors. Although these methods are computationally more expensive, recent progress in optimization techniques and computer hardware have made them feasible for realistic robotic systems. Some representative methods along this direction include Linear Quadratic Regulator (LQR) \cite{Posa2016}, Model Predictive Control (MPC) \cite{Erez2013}, \cite{Koenemann2015}, and Hybrid Zero Dynamics (HZD) \cite{Westervelt2007}, \cite{Hereid2018}. In particular, HZD is a formal framework for the control of bipedal robots with or without underactuation through the design of nonlinear feedback controllers and a set of virtual constraints. It has been successfully implemented in several physical robots, including many underactuated robots \cite{Chevallereau2003,Chevallereau2009,Sreenath2011,Hereid2014,Hereid2018}.

Most of the aforementioned results are based on analytic models of the robot, which can be hard to derive for complex robotic systems. In addition, biped walking dynamics include contact and collision between the robot and the ground, which makes precise modeling of the dynamics difficult. 
Recently, there is an increasing interest in using Reinforcement Learning (RL) to obtain effective control policies using the dynamic simulation of the robot. Some early approaches use RL with Central Pattern Generator (CPG)-based controllers using the cerebellar model arithmetic computer (CMAC) neural networks \cite{Benbrahim1997}. More recent work use state-of-the-art policy gradient methods to find policies that map from the observation space to the action space in order to achieve a continuous walking motion \cite{Lillicrap2015}, \cite{Schulman2015}. However, general RL methods combined with deep neural networks can be sampling inefficient (millions of data samples) and are usually over-parameterized (thousands of tunable parameters) as they do not consider the underlying physics of bipedal walking. This may lead to unnatural motions that are not applicable to real robots. In addition, to our best knowledge existing model-free RL methods in the literature consider neither regulating the walking speed of the robot nor the local stability of the walking gaits.

Some efforts to address the velocity regulating problem rely on the use of Supervised Learning (SL) as a tool for obtaining a policy that renders stable dynamic walking gaits for different speeds \cite{Da2016, Da2017}. The authors proposed an offline approach to design an explicit model-based feedback control policy based on HZD and SL. However, these methods still require the knowledge of an analytic model of the robot.

This paper focuses on developing a novel model-free RL approach for bipedal walking control that employs a non-traditional RL structure with an embedded HZD framework. The proposed method does not need an analytic form of the robot's model. Instead, it uses a realistic physics simulator that can capture the interactions between the robot and the environment. The proposed structure of the neural network used for the training of the control policy does not use the full state space of the robot, but a reduced state space. By means of the HZD, the outputs of the neural network are mapped into a set of polynomials that define the desired outputs for the actuated joints of the robot. This allows reducing the number of parameters of the neural network. Then, the desired outputs are tracked by an adaptive PD controller, which ensure the compliance of the HZD virtual constraints. 

The main contribution of the paper is on improving the existing RL methods for training bipedal walking control by incorporating some key insights from the HZD into the learning process. We believe that incorporating physical insights of bipedal walking can significantly improve the training results of the RL and make them more realistic and applicable to real robots. To the best of our knowledge, this is the first time HZD is combined with RL to realize feedback controllers for bipedal walking. The result of combining RL and HZD allows the learned control policy to track different desired speeds within a continuous interval. Moreover, the learned controller outperforms the traditional HZD-based controller regarding robustness while still maintaining the stability of the walking limit cycle, which is one of the key features of the HZD approach.

Finally, we demonstrate the feasibility and effectiveness of the proposed method by evaluating the performance of the learned policy for speed tracking and robustness to external disturbances on the simulation of a five link-planar underactuated robot on MuJoCo -a novel physics engine \cite{Todorov2012}. In this paper, the model of RABBIT robot is used; however, the method can be extended to other robot models.

\section{PROBLEM FORMULATION}
In this section, we present the description of the robot used for the implementation and evaluation of the proposed method. Moreover, we provide some backgrounds for HZD and RL, which are the main components of the proposed control strategy.

\subsection{Robot Description}
\begin{figure}[h]
\centering
\includegraphics[width=3.5cm]{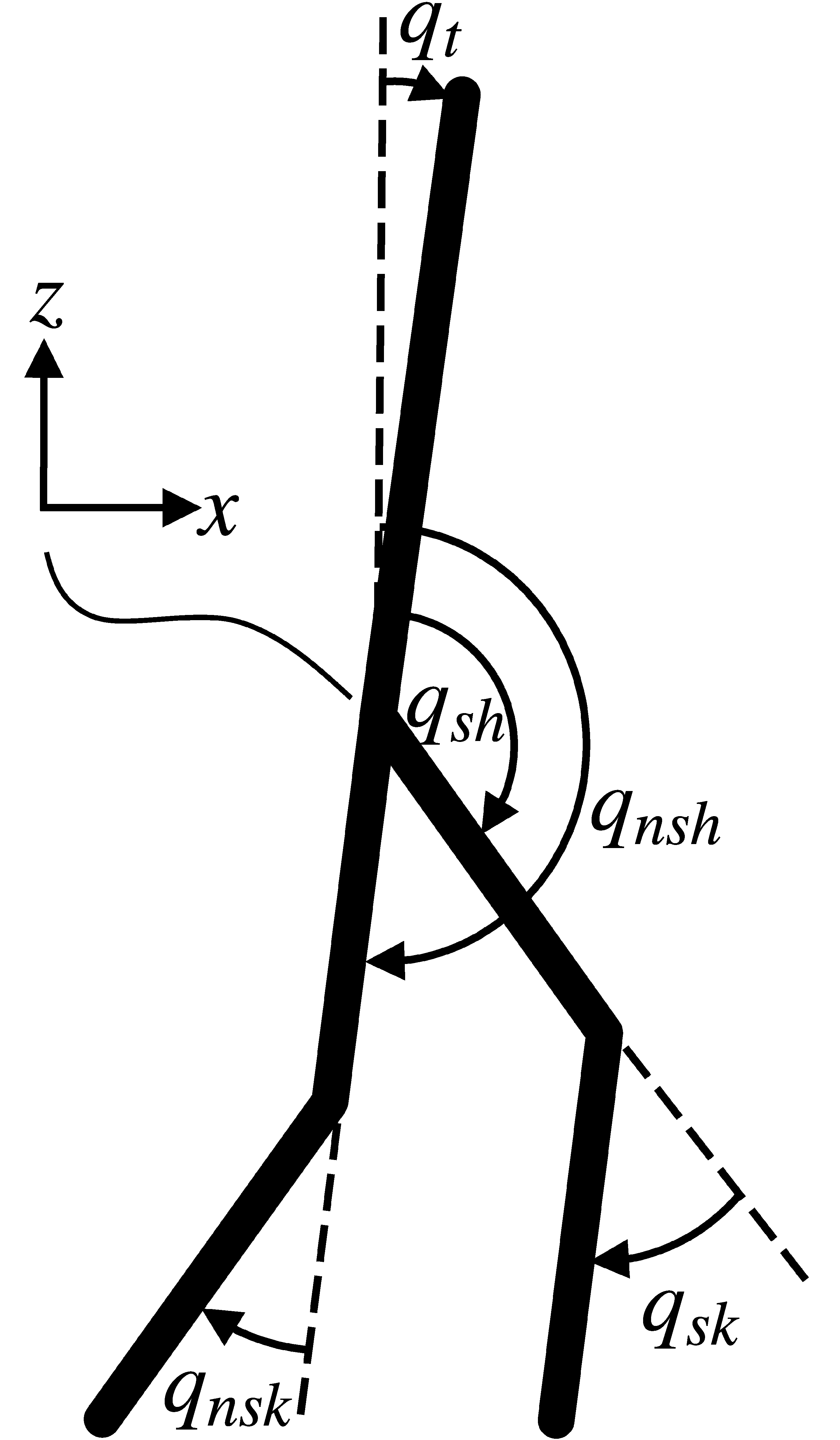}
\caption{Coordinate system schematic of RABBIT robot}
\label{fig:Rabbit_model}
\end{figure}

As a starting point for the proposed method, we consider the model of the robot RABBIT, which is a well known test-bed robot model for the HZD framework \cite{Chevallereau2003}. Despite its simple mechanical structure, RABBIT is still a good representation of biped locomotion. RABBIT is a five-link, planar underactuated bipedal robot with a total weight of 32 kg. The five links of the robot correspond to the torso, right thigh, right shin, left thigh and left shin. The robot has point feet and four actuated joints, two in the hip joints and two in the knee joints. This configuration results in a five degree-of-freedom mechanism during the single support phase (considering the stance leg end does not slip) and four degrees of actuation. In the upright position, with both legs together and straight, the hip is 80 cm above the ground, and the tip of the torso is at 1.43 m \cite{Westervelt2007}. See Table \ref{table_1} for a description of the length, mass, and inertia of each link of the robot. All these parameters have been included in the simulation model used for both the training process and the evaluation of the learned control policy. Fig. \ref{fig:Rabbit_model} shows the schematic of RABBIT and the notation used in this paper for the description of the state variables of the robot.

\begin{table}[h!]
\centering
\begin{tabular}{|l|l|l|l|l}
\cline{1-4}
                            & Torso & Femur & Tibia &  \\ \cline{1-4}
Length $[m]$                & 0.63  & 0.4   & 0.4   &  \\ \cline{1-4}
Mass $[kg]$                 & 12    & 6.8   & 3.2   &  \\ \cline{1-4}
Inertia $[kg \cdot m^{2}]$  & 1.33  & 0.47  & 0.2   &  \\ \cline{1-4}
\end{tabular}
\caption{Model parameters of RABBIT robot}
\label{table_1}
\end{table}
\vspace*{-5mm}

\vspace*{-4mm}

\subsection{Hybrid Zero Dynamics}

We now briefly review some key concepts and ideas about HZD that are useful for developing our new reinforcement learning framework. In the HZD based controllers, virtual constraints are introduced as a means to synthesize feedback controllers that realize stable and dynamic locomotion. By designing virtual constraints that are invariant through impact, an invariant sub-manifold is created---termed the \emph{hybrid zero dynamics surface}---wherein the evolution of the system is dictated by the dynamics of the reduced-dimensional underactuated degrees of freedom of the system \cite{Westervelt2007, Ames2013Human}. 

Let $q = (q_t, q_{sh}, q_{sk}, q_{nsh}, q_{nsk})$ be the joint coordinates of RABBIT (see Fig.~\ref{fig:Rabbit_model}) and $\tau(q) \in [0,1] $ be a state-based time representation (see \eqref{eq:tau} for explicit definition), then virtual constraints are defined as the difference between the actual and desired outputs of the robot \cite{Ames2013Human}:
\begin{align}
  y_{2} &:= y^a_{2}(q) - y^d_{2}(\tau(q),\alpha),
  \label{eq_vc}
\end{align}
where $y_{2}$ is (vector) relative degree 2 due to the second order dynamical system of the robot mechanical model, and $y^d_{2}$ is a vector of desired outputs defined in terms of $5^{\text{th}}$ order B\'ezier polynomials parameterized by the coefficients $\alpha$, given as:
\begin{align}
    y^d_2(\tau(q),\alpha) := \sum_{k=0}^{5} \alpha[k] \frac{M!}{k!(M-k)!} \tau(q)^k (1-\tau(q))^{M-k}.
\end{align}
Moreover, the B\'ezier polynomial has a very nice feature: $y^d_2(0,\alpha) = \alpha[0]$ and $y^d_2(1,\alpha) = \alpha[5]$. This will be used later to reduce the search space of the neural network parameters in the proposed RL approach.

In this paper, we choose $\tau(q)$ to be the scaled relative forward hip position with respect to the stance foot, i.e.,
\begin{align}
    \label{eq:tau}
    \tau(q) = \frac{p_{hip}(q) - p_{hip}^-}{p_{hip}^+ - p_{hip}^-}
\end{align}
where $p_{hip}^+$ and $p_{hip}^-$ are the values of $p_{hip}(q)$ at the beginning and end of a step. It can be noted that by driving virtual constraints to zeros through feedback controllers, the trajectories of all joints are synchronized to the evolution of the relative forward hip motion, i.e., the zero dynamics of the system. By properly choosing the coefficients of these B\'ezier polynomials, one can achieve different walking motions. More importantly, the local stability of the periodic walking gait can be formally validated by computing the Poincar\'e map of the reduced dimensional zero dynamics surface.

\subsection{Reinforcement Learning}

Generally speaking, RL aims to find an approximate solution to an optimal control problem of a certain class of dynamical systems, which can be formalized as follows:
\begin{align}
    \label{eq:optimal-control-problem}
    & \underset{x}{\text{maximize}}
    & & J(\theta) = E_{p((s_t, a_t);\theta)} \left[ \sum_{t=1}^{T} \gamma^t r(s_t, a_t) \right], \gamma \in (0, 1], \nonumber\\
    & \text{subject to}
    & & s_{t+1} = f(s_t, a_t).
\end{align}
That is, given state $s_t$ with dynamics transition $f$, one seeks to maximize the expected discounted accumulated reward $r(s_t, a_t)$ through the action sequence $a_t$. The trajectory distribution $p((s_t, a_t);\theta)$ is induced by the probabilistic policy $a_t \sim \pi(s_t | \theta)$. With policy being a neural network, $\theta$ represents the set of network parameters.

The main approaches to solve RL problems are based on either value iteration \cite{Mnih2013} or policy gradient \cite{Lillicrap2015}, \cite{Schulman2017}. Value iteration, such as Q-learning \cite{Mnih2013}, takes advantage of the recursive form of the Bellman equation to establish an off-policy algorithm to learn the action-value function $Q^{\pi}(s_t, a_t)$. While such methods have shown promising performance on complex tasks with high-dimensional state space, it can only handle discrete, low-dimensional action space. For robotic applications with continuous high-dimensional action space, policy gradient methods are more commonly adopted \cite{Lillicrap2015}.

In this paper, we adopted two state-of-the-art RL algorithms in our simulation, including Evolution Strategies (ES) \cite{Gomez2005} and Proximal Policy Gradient (PPO) \cite{Schulman2017}. Both methods estimate the policy gradient
\begin{align}
    \label{eq:policy gradient}
    \triangledown_{\theta} J(\theta) = E_{p(x_t;\theta)}\left[ r(x_t) \triangledown_{\theta} \log p(x_t;\theta) \right], x_t = (s_t, a_t),
\end{align}
either implicitly (ES) or explicitly (PPO). The policy is iteratively improved through simulation rollouts followed by gradient accent with respect to the objective. ES is one of the random search methods where a population of $N$ policies ${\pi(s|\theta_i)}_{i=1}^{N}$ are sampled following $\theta_i \sim \mathcal{N}(\theta_{\mu}, \sigma)$. The normal distribution of policy parameterized on $\theta_{\mu}$ and $\sigma$ is then improved through estimated gradient using the evaluation results from the sampled policies. PPO proposes a novel objective function of
\begin{align}
    \label{eq:clip loss}
    L^{CLIP}(\theta) = E_t \left[ \min(g_t(\theta)A_t, clip (g_t(\theta), 1-\epsilon, 1+\epsilon)A_t) \right],
\end{align}
where $A_t$ is the so-called advantage estimation \cite{schulman2015high}. The policy $\pi_{\theta}(a_t|s_t)$ is improved by a modified probability ratio of $g_t(\theta)=\frac{\pi_{\theta}(a_t|s_t)}{\pi_{\theta_{old}}(a_t|s_t)}$ controlled by the clip ratio $\epsilon$. 

In our simulation, both training methods provide similar results in terms of training speed, sampling efficiency, and policy performance. They are also sharing the same neural network structure (see section III). We will not distinguish between these two methods in later sections.

\section{HYBRID ZERO DYNAMICS BASED REINFORCEMENT LEARNING (HZD-RL)}

This section will introduce the proposed control-learning framework that combines HZD-based control design with reinforcement learning. We will first lay out the overall control-learning structure and then provide technical details for some key components in the framework. 

\subsection{Control-Learning Structure}
Traditional RL algorithms for bipedal walking search for control policies that directly map the current robot's states to the control action. We propose a non-traditional structure for the RL framework, whose resulting control policy maps from a reduced order of the robot's state to a set of coefficients of the B\'ezier polynomials that define the trajectory of the actuated joints. It is worth noticing that general RL algorithms adopt probabilistic policies to facilitate the training process. As a result, the system trajectories become stochastic despite we start with a deterministic robot model. In addition, as the desired policy needs to be able to perform speed tracking, we also consider the desired velocity and the velocity tracking error. Finally, we use an adaptive low level PD controller for tracking the desired output for each joint. This enforces the compliance of the HZD virtual constraints, which render stable and robust locomotion for the bipedal robot. It is worth mentioning that unlike some RL methods, reference trajectories for the robot's joints are not given as an input of the RL process in our approach \cite{Levine2013, Peng2017, Xie2018}. Instead, they are naturally obtained by the proposed HZD-RL structure.

\begin{figure}[t]
\vspace{2mm}
\centering
\includegraphics[width=8.5cm]{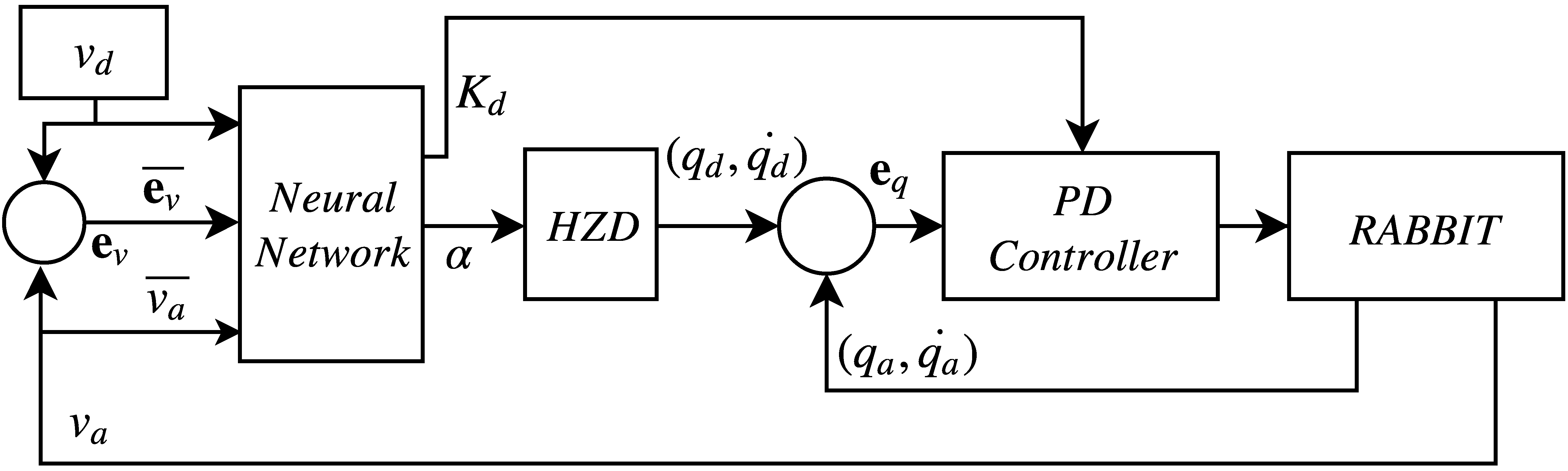}
\caption{Learning-Control Structure}
\label{fig:CL_struct}
\vspace{-4mm}
\end{figure}

Fig. \ref{fig:CL_struct} presents a diagram of the overall control-learning structure. For each time step, the desired walking speed ($v_d$) and the actual speed of the robot's hip ($v_a$) are used to generate the inputs of the neural network. A Detailed explanation of the neural network structure will be provided in the subsection B. The trained control policy maps directly the inputs of the neural network to the set of coefficients $\alpha$ and the controller's derivative gain $K_d$. Then, $\alpha$ jointly with the phase variable $\tau$ are used to compute the desired joint's position and velocity for each actuated joint of the robot by means of the HZD framework. The adaptive PD controller uses the tracking error between the desired and actual value of the output to compute the torque of each actuated joint, which is the input of the dynamic system that represents the walking motion of the robot. Finally, the measured outputs of the system (states of the robot) are used as feedback for the inner and outer control loops.

As part of the RL framework, we need to establish the reward function that will be used during the training process. This reward function is defined in the quadratic form
\begin{align}
r(s_t, a_t) = |\overline{v_a}-v_d|^2
\end{align} 
with constraints conditions 
\begin{align}
    |q_t|<0.5, \quad 0.6<z<0.8, 
\end{align}
and steps of rollouts with discount factor $\gamma=0.99$.

\subsection{Neural Network Structure}
\begin{figure}[t]
\vspace{2mm}
\centering
\includegraphics[width=8.5cm]{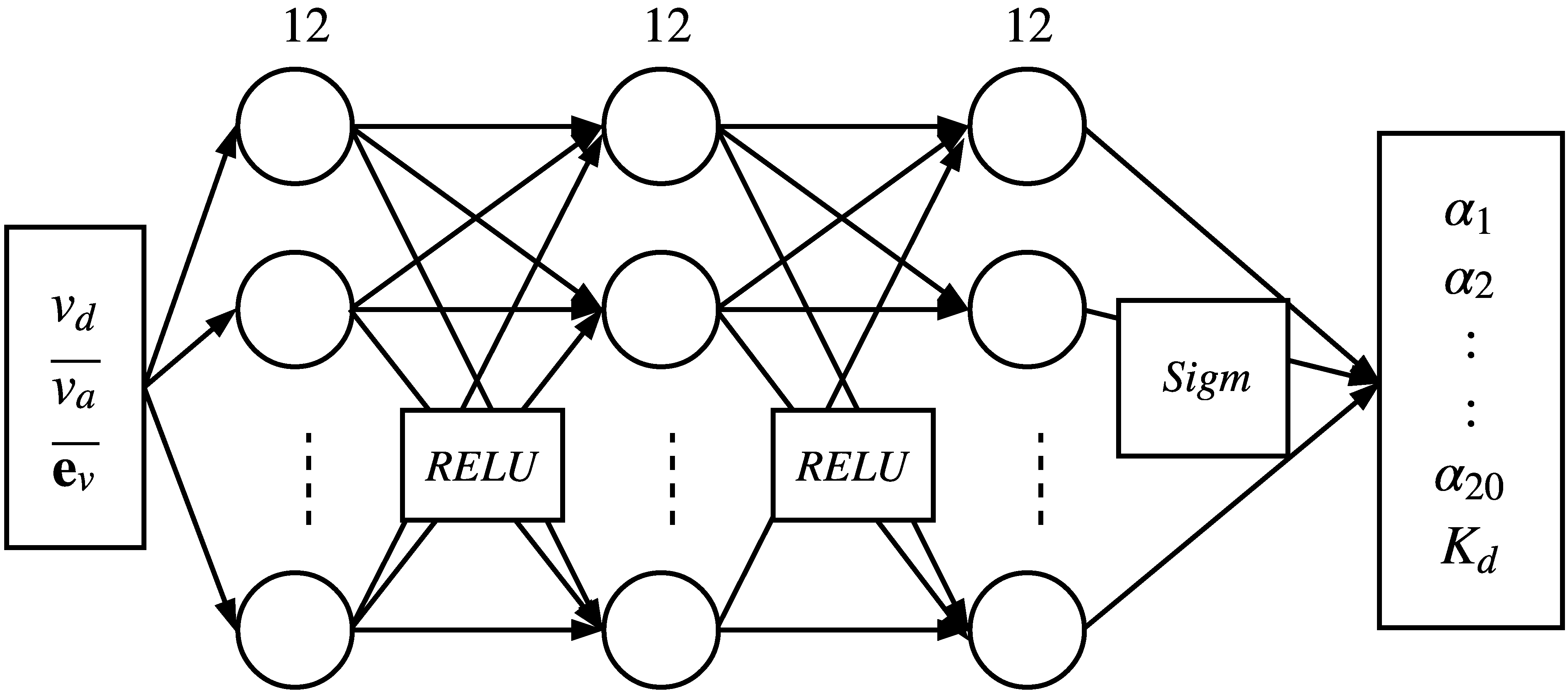}
\caption{Structure of the Neural Network}
\label{fig:NN_struct}
\vspace*{-4mm}
\end{figure}

The structure of the neural network used during the training process is presented in Fig. \ref{fig:NN_struct}. Because of the complex dynamics of the walking motion, it is impossible to guarantee a good tracking performance for the instantaneous speed of the robot along the $x$ axis, which corresponds to the walking direction of the robot. Therefore, it is necessary to consider the average speed of the walking motion. Here, we consider this average speed to be the speed during about one walking step of the robot, which takes about 200 simulation steps. Therefore, the inputs of the neural network are the value of the desired velocity ($v_d$), the average hip's velocity ($\overline{v_a}$) of the robot for the last 200 simulation steps, and the average error between the desired velocity and the instantaneous velocity of the robot's hip during the last 200 simulation steps. The value of the desired velocity is uniformly sampled from a continuous space interval from 0.7 to 1.5 $m/s$.

The output of the neural network corresponds to the values for the coefficients of the B\'ezier polynomials. Since the robot has four actuated joints and each B\'ezier polynomial is of degree 5, the total size of the set of parameters $\alpha$ is 24. However, to encourage the invariance through impact of the virtual constraints, we enforce the position of the hip joints ($q_{sh}$, $q_{nsh}$) and knee joints ($q_{sk}$, $q_{nsk}$) to be equal at the beginning and end of the step ($\tau(q)=0$ and $\tau(q)=1$ respectively). This leads to the following set of equalities. 
\begin{align}
  \alpha[1] = \alpha[23]; \alpha[2] = \alpha[24]; \alpha[3] = \alpha[21];   \alpha[4] = \alpha[22] \nonumber
\end{align}
Therefore, the number of outputs of the neural network is reduced to 20. Additionally, we consider as an output of the neural network the derivative constant of the PD controller used for the tracking of the desired outputs. The number of hidden layers is 3, each one with 12 neurons, and the final layer employs a sigmoid function to limit the range of the outputs. 
A very important feature of the proposed HZD-RL structure is the physical insight that the set of coefficients of the B\'ezier polynomials have. Due to the family of polynomials chosen to construct the desired outputs, this set of parameters defines the waypoints for the trajectories of the desired outputs. Therefore, the output range of the set of parameters can be limited by the physical constraint of each actuated joints. This important feature allows reducing the continuous interval of the output, which greatly decreases the complexity of the RL problem and improves the efficiency and effectiveness of the learning process. 

\subsection{Adaptive PD controller}
Adaptive control can improve the performance of a control law to handle the uncertainty of unknown parameters in a system and the complexity of highly non-linear models. Early approaches in adaptive control discuss the advantages of using the tracking error as an update law for adaptive controllers based on adaptive inverse dynamics control \cite{Ortega1988}, \cite{Slotine1989}. More recent work combines the idea of adaptive control with machine learning to obtain controllers whose parameters evolve with the dynamics of the process \cite{Wang2007}, \cite{Manoonpong2007}. In this paper, we use an adaptive PD controller with fixed proportional gain and variable derivative gain to compute the torque applied to each actuated joint. For this, each controller takes the position tracking error and the velocity tracking error of each joint according to the following structure.
\begin{align}
  u &= K_p\textbf{e} + K_d\dot{\textbf{e}}
\end{align}  
with $\textbf{e} = q_d  - q_a$ and $\dot{\textbf{e}} = \dot{q}_d - \dot{q}_a$,
where $q_d$ and $q_a$ are the desired and actual joint positions, $\dot{q}_d$ and $\dot{q}_a$ are the desired and actual joint velocities, respectively.

It is important to clearly state the strong connection between the adaptive PD controller, the HZD framework, and the RL structure. From section II, we know that HZD virtual constraints are defined by equation \ref{eq_vc}, where the desired output is defined in terms of $5^{th}$ order B\'ezier polynomials. Therefore, the desired position and velocity for each joint are defined by 
\begin{align}
    q_d &= y_d(\tau(q), \alpha) \\
    \dot{q}_d &= \dot{y}_d(\tau(q), \alpha)
\end{align}
Since the set of coefficients $\alpha$, and the controller's derivative constant are outputs of the neural network, we can compute them by using the learned control policy resulting from the training process. This is, $[\alpha,K_d]={\pi(s|\theta)}$.

Finally, at each simulation step, the output of each PD adaptive controller (torque of each actuated joint) results in an output of the dynamic system (states of the robot), which is used as feedback for both the adaptive PD controller and the neural network. This closes the loop of the feedback control process. The described controller can adapt its behavior to the changes in the dynamics of the walking cycle and the disturbances inherent from the hybrid dynamics and external forces. Particularly, such adaptability feature turns out to be highly useful for speed tracking when a change of the desired speed is detected and for disturbance rejection.

\section{SIMULATION RESULTS}
The implementation of the customized environment for RABBIT was build in OpenAI Gym \cite{Brockman2016}, and the environment was simulated using the MuJoCo physics engine \cite{Todorov2012}. The number of trainable parameter for the neural network is 620, and the training time is about 30 minutes using a 12-core CPU machine. Visualized results of the learning process and evaluation of the policy in simulation can be seen in the accompanying video submission (can also be found in \cite{video_link}).

In order to encourage a good speed tracking performance on the learned policy, for each episode of rollout the desired velocity is updated once in the same way it is chosen at the beginning of the episode (uniform sampling).
The control policy obtained from the training is evaluated for several scenarios including speed tracking and convergence of the walking limit cycle. Finally, we evaluated the robustness of the learned policy using the HZD-RL method and compared it with the control policy obtained using the traditional model-based HZD framework. 

\subsection{Speed tracking}
\begin{figure}[h]
\vspace{2mm}
\centering
\includegraphics[width=1.0\columnwidth]{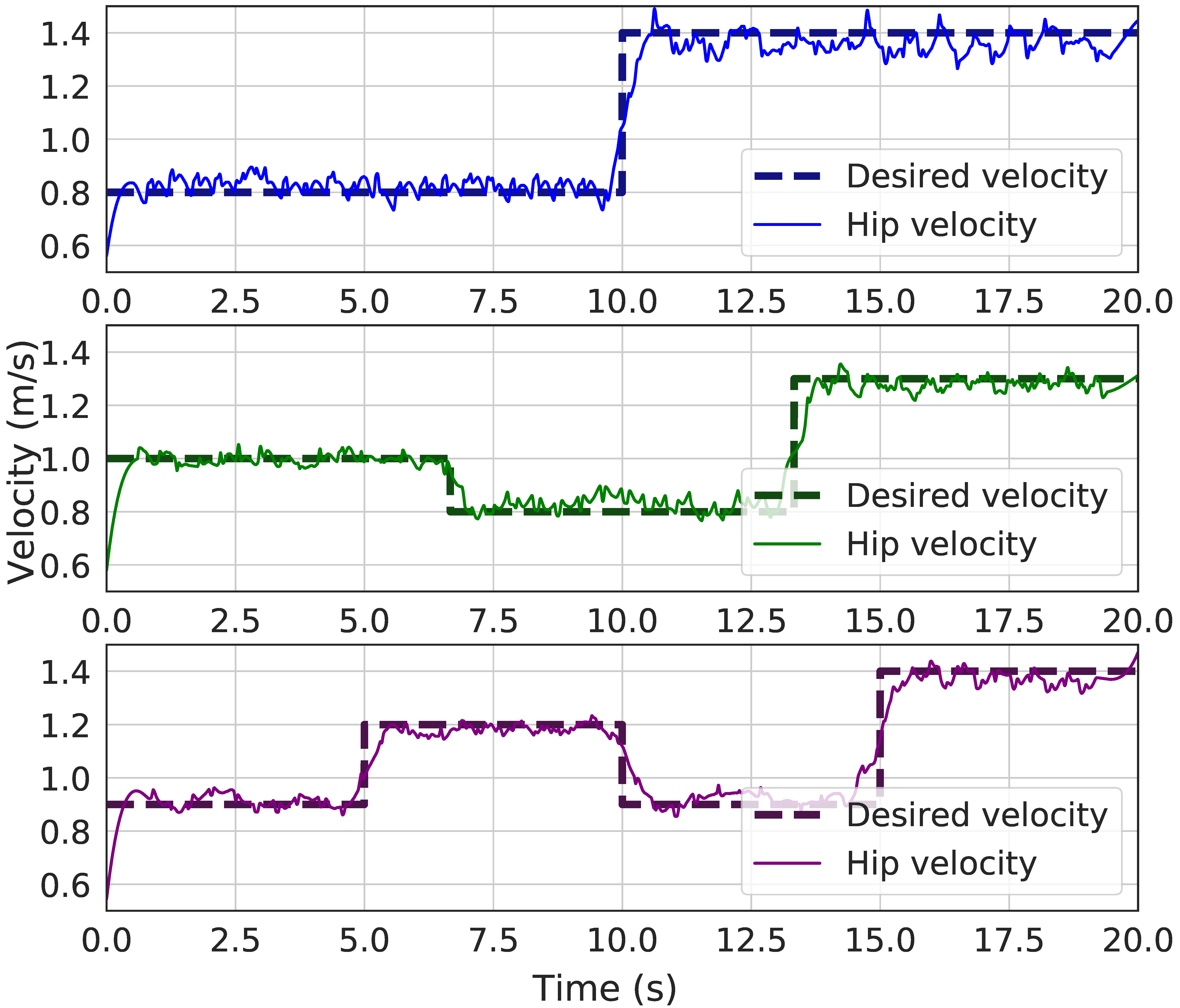}
\caption{Speed tracking performance of the learned policy}
\label{change_vel}
\end{figure}

The learned policy was tested for tracking one specific desired speed in several scenarios, including tracking a fixed desired speed and a range of variable desired speeds. Fig. \ref{change_vel} shows the filtered instantaneous speed when the robot walks while tracking a set of different desired speeds. The policy's performance is good, and it allows the robot to track effectively the fixed speeds as well as speed changes. The plot only shows the speed tracking results for 20 seconds; however, since the policy renders a stable walking limit cycle, the robot is able to keep walking for much longer time. This aspect is discussed further in the next subsection.

\subsection{Stability of the  walking limit cycle}
\begin{figure}[h]
\centering
\includegraphics[width=1\columnwidth]{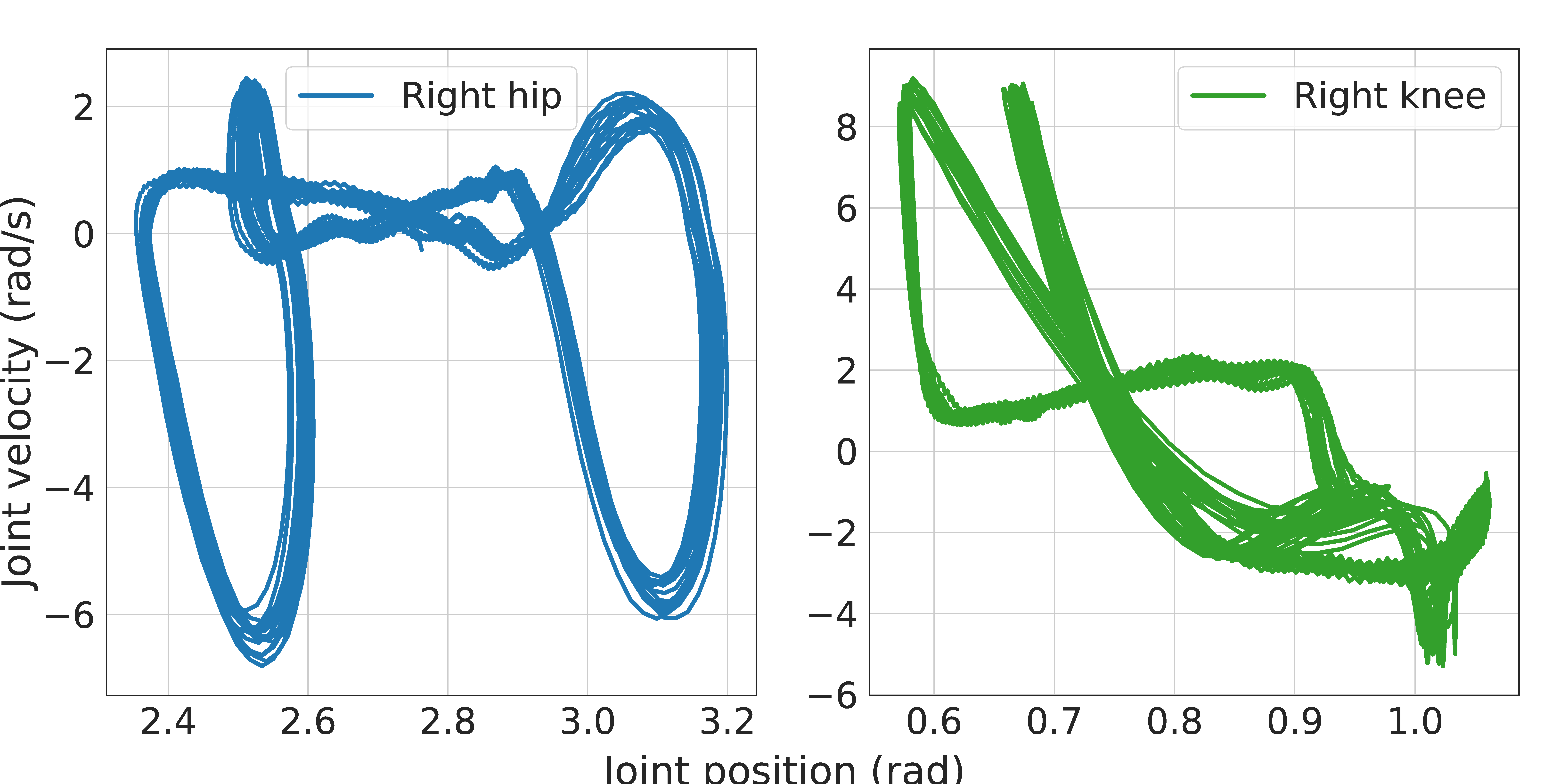}
\caption{Walking limit cycle of the learned policy}
\label{limit_cycle}
\end{figure}

One of the main advantages of the HZD is that it provides a formal framework to prove the stability of the walking limit cycle \cite{Plestan2003}. Therefore, since HZD is the underlying layer of our proposed method, we analyze not only the effectiveness of the policy for tracking a desired fixed or variable speed, but also the convergence of the walking limit cycle. Fig. \ref{limit_cycle} presents the limit cycle over several steps with the parameters defined by the learned policy. The resulting trajectory converges to a limit cycle, supporting the stability analysis presented in the HZD theory \cite{Westervelt2007}, \cite{Chevallereau2009}, \cite{Plestan2003}.

\vspace*{-3mm}

\subsection{Disturbance rejection and robustness comparison}
To evaluate the robustness of the learned control policy, we applied an external force directly into the robot's torso in three different scenarios: 1) Small forces applied in the forward direction. 2) Small forces applied in the backward direction. 3) Large forces applied to the backward direction.

The controller used for the comparison test is obtained from the Fast Robot Optimization and Simulation Toolkit (FROST), which is a software environment for developing model-based control for robotic systems using the HZD framework \cite{Hereid2017}. This model-based controller is implemented in the environment simulation of RABBIT under the same conditions used for the evaluation of the HZD-RL policy.

\begin{figure}[!ht]
\vspace*{2mm}
\subfloat[Speed tracking during external disturbance]{%
  \includegraphics[clip,width=1\columnwidth]{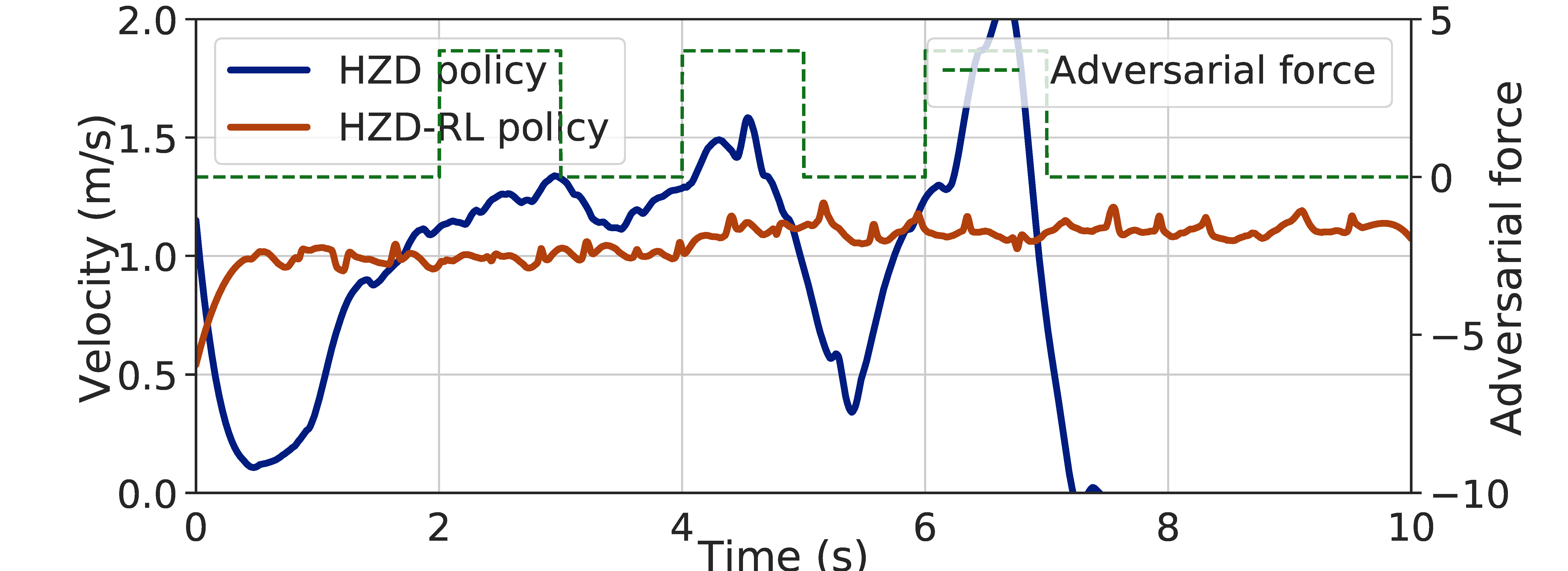}
  \label{comp_small_fw_a}%
}

\subfloat[Motion produced by HZD policy]{%
  \includegraphics[clip,width=1\columnwidth]{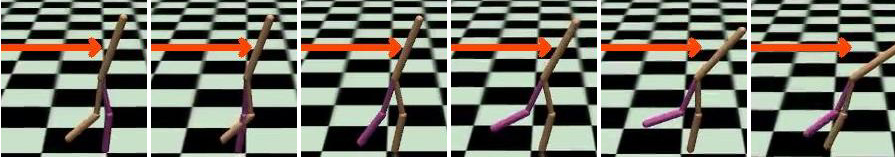}
  \label{comp_small_fw_b}%
}

\subfloat[Motion produced by HZD-RL policy]{%
  \includegraphics[clip,width=1\columnwidth]{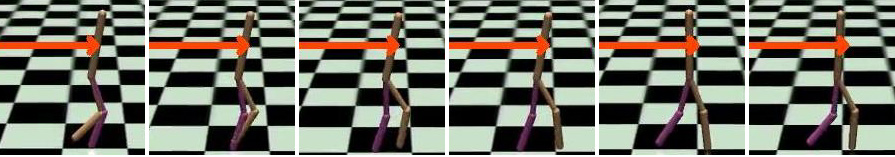}
  \label{comp_small_fw_c}%
}

\caption{Robustness comparison between HZD-RL and HZD when a small force is applied in the forward direction.}
\label{comp_small_fw}
\end{figure}

\begin{figure}[!ht]
\vspace{2mm}
\subfloat[Speed tracking during external disturbance]{%
  \includegraphics[clip,width=1\columnwidth]{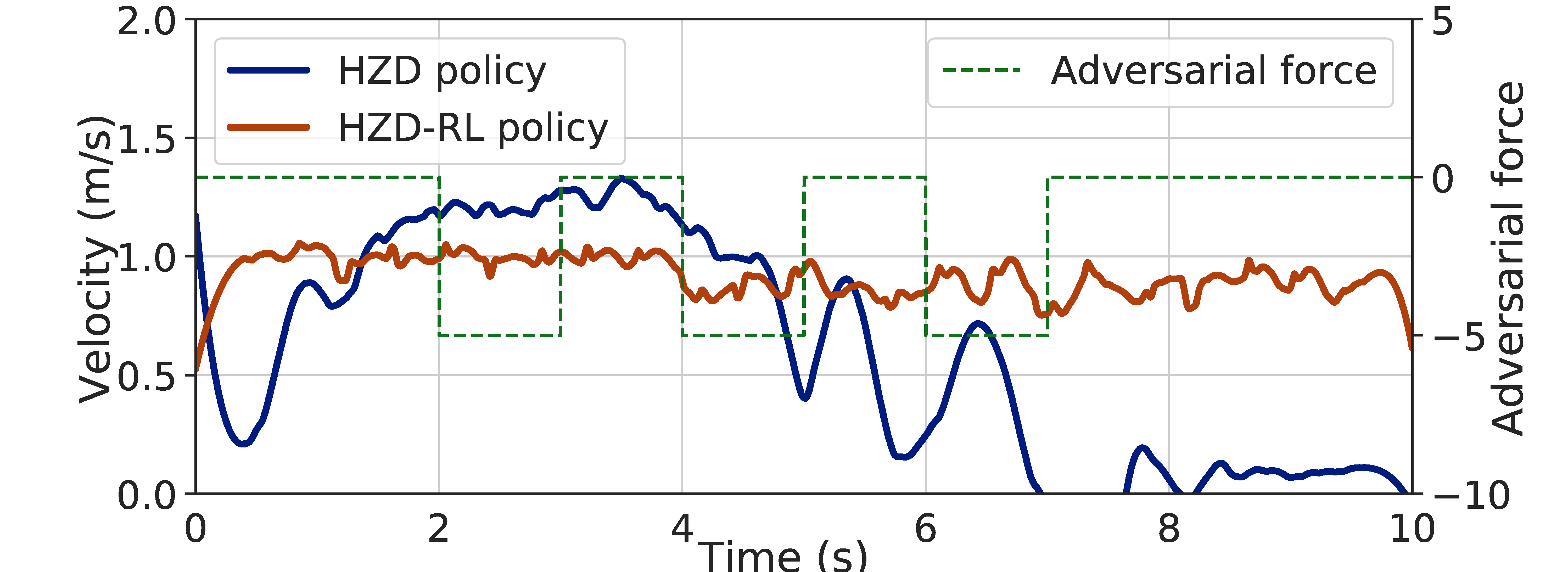}
  \label{comp_small_bw_a}%
}

\subfloat[Motion produced by HZD policy]{%
  \includegraphics[clip,width=1\columnwidth]{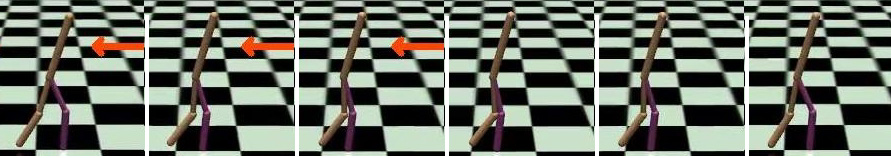}
  \label{comp_small_bw_b}%
}

\subfloat[Motion produced by HZD-RL policy]{%
  \includegraphics[clip,width=1\columnwidth]{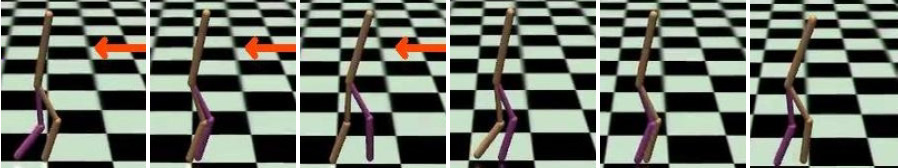}
  \label{comp_small_bw_c}%
}

\caption{Robustness comparison between HZD-RL and HZD when a small force is applied in the backward direction.}
\label{comp_small_bw}
\end{figure}

\begin{figure}[!h]
\vspace{2mm}
\subfloat[Speed tracking during external disturbance]{%
  \includegraphics[clip,width=1\columnwidth]{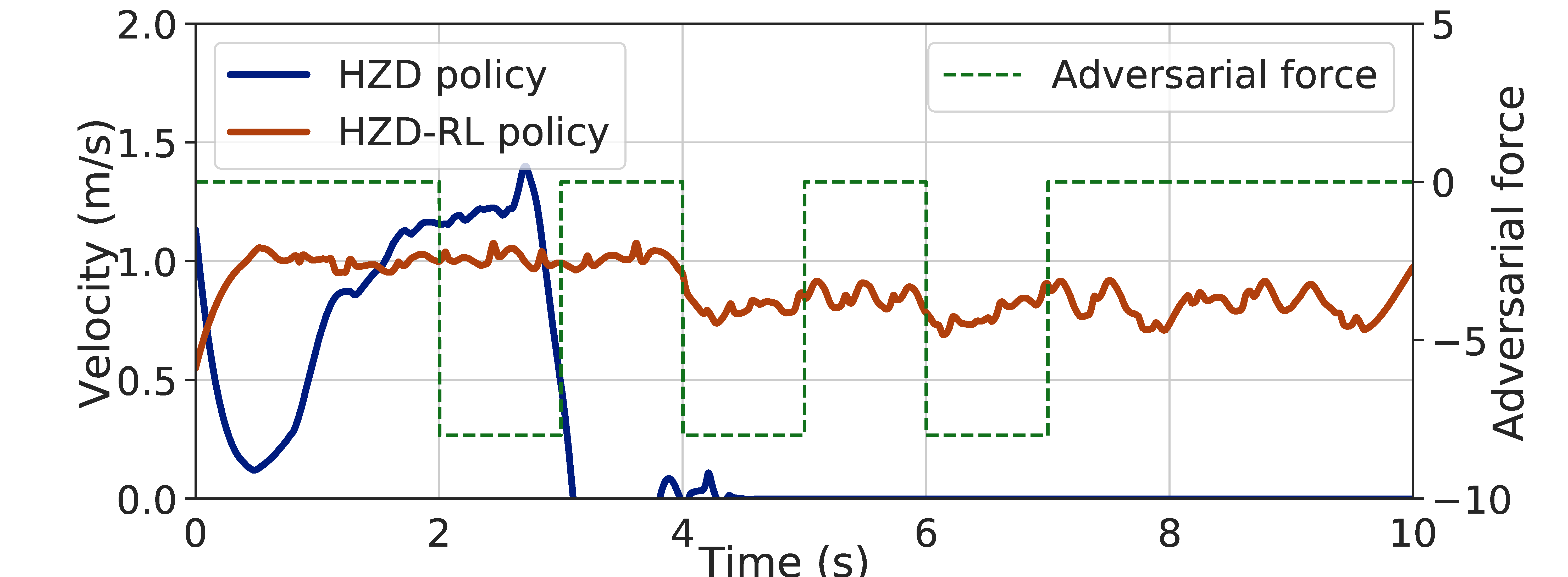}
  \label{comp_large_bw_a}%
}

\subfloat[Motion produced by HZD policy]{%
  \includegraphics[clip,width=1\columnwidth]{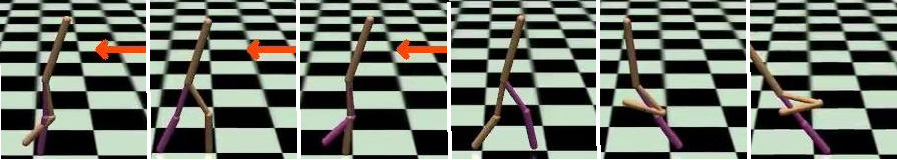}
  \label{comp_large_bw_b}%
}

\subfloat[Motion produced by HZD-RL policy]{%
  \includegraphics[clip,width=1\columnwidth]{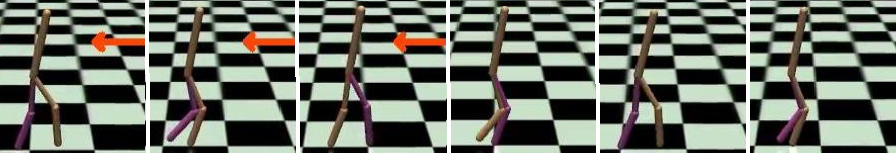}
  \label{comp_large_bw_c}%
}
\caption{Robustness comparison between HZD-RL and HZD when a large force is applied in the backward direction.}
\label{comp_large_bw}
\end{figure}
Fig. \ref{comp_small_fw} illustrates the comparison of the policies for case 1. A small external force is applied in the forward direction for both controllers at time $t=2s$, $t=4s$, and $t=6s$ using the same technique described in \cite{Pinto2017}. Fig. \ref{comp_small_fw_a} shows the response of both controllers. The traditional HZD controller cannot reject the disturbance and the robot falls while the HZD-RL control policy is able to recover from the disturbance and allows the robot to continue walking. Visual details of the performance of the HZD and HZD-RL policies are shown in Fig. \ref{comp_small_fw_b} and \ref{comp_small_fw_c} respectively. 

Fig. \ref{comp_small_bw} shows the result of the comparison in case 2. When the external force is applied, the HZD controller cannot finish the step, goes back to the start position of that step, and get stuck there. This effect can be appreciated in Fig. \ref{comp_small_bw_b}, where the speed decreases gradually to $0\:(m/s)$. On the other hand, the HZD-RL policy permits the robot to complete the step and recover from the external disturbance while maintaining tracking the desired walking speed. Fig. \ref{comp_small_bw_b} and \ref{comp_small_bw_c} provide visual details of the consequent walking motion. 

Finally, Fig. \ref{comp_large_bw} shows the comparison for case 3. Fig. \ref{comp_large_bw_a} shows the abrupt speed change in the robot speed caused by the robot falling to the ground when the HZD controller is used. The HZD-RL controller keeps the robot walking after the external force is applied. Visual details of the performance of both controllers is shown in
Fig. \ref{comp_large_bw_b} and \ref{comp_large_bw_c}.

\section{CONCLUSIONS}
This paper presents a novel model-free RL approach for the design of feedback controllers based on the HZD framework. We showed that by considering the physics insight of the bipedal walking into the structure of the RL, we can obtain a control policy that is able to track different walking speeds within a continuous interval. The proposed structure is simple, leading to reduced the number of parameters of the neural network. The proposed method is validated through simulation of the RABBIT robot, where the stability and robustness of the learned policy is evaluated. The results show a good performance of the learned policy for tracking any desired speed within a continuous range while maintaining stability of the walking limit cycle. Finally, the robustness comparison showed that the learned control policy outperforms the traditional model-based HZD controller when recovering from external disturbances.







\newpage

\begin{thebibliography}{10}
\providecommand{\url}[1]{#1}
\csname url@rmstyle\endcsname
\providecommand{\newblock}{\relax}
\providecommand{\bibinfo}[2]{#2}
\providecommand\BIBentrySTDinterwordspacing{\spaceskip=0pt\relax}
\providecommand\BIBentryALTinterwordstretchfactor{4}
\providecommand\BIBentryALTinterwordspacing{\spaceskip=\fontdimen2\font plus
\BIBentryALTinterwordstretchfactor\fontdimen3\font minus
  \fontdimen4\font\relax}
\providecommand\BIBforeignlanguage[2]{{%
\expandafter\ifx\csname l@#1\endcsname\relax
\typeout{** WARNING: IEEEtran.bst: No hyphenation pattern has been}%
\typeout{** loaded for the language `#1'. Using the pattern for}%
\typeout{** the default language instead.}%
\else
\language=\csname l@#1\endcsname
\fi
#2}}

\bibitem{Kajita1992}
S.~Kajita, T.~Yamaura, and A.~Kobayashi, ``Dynamic walking control of a biped
  robot along a potential energy conserving orbit,'' \emph{IEEE Transactions on
  Robotics and Automation}, vol.~8, no.~4, pp. 431--438, Aug 1992.

\bibitem{Vukobratovic2004}
M.~Vukobratovic and B.~Borovac, ``Zero-moment point - thirty five years of its
  life,'' \emph{I. J. Humanoid Robotics}, vol.~2, pp. 225--227, 2004.

\bibitem{Yoshida2008}
E.~Yoshida, C.~Esteves, I.~Belousov, J.~Laumond, T.~Sakaguchi, and K.~Yokoi,
  ``Planning 3-d collision-free dynamic robotic motion through iterative
  reshaping,'' \emph{IEEE Transactions on Robotics}, vol.~24, no.~5, pp.
  1186--1198, Oct 2008.

\bibitem{Stephens2010}
B.~J. Stephens and C.~G. Atkeson, ``Dynamic balance force control for compliant
  humanoid robots,'' in \emph{2010 IEEE/RSJ International Conference on
  Intelligent Robots and Systems}, Oct 2010, pp. 1248--1255.

\bibitem{Pratt2006}
J.~Pratt, J.~Carff, S.~Drakunov, and A.~Goswami, ``Capture point: A step toward
  humanoid push recovery,'' in \emph{2006 6th IEEE-RAS International Conference
  on Humanoid Robots}, Dec 2006, pp. 200--207.

\bibitem{Pratt2012}
J.~Pratt, T.~Koolen, T.~de~Boer, J.~Rebula, S.~Cotton, J.~Carff, M.~Johnson,
  and P.~Neuhaus, ``Capturability-based analysis and control of legged
  locomotion, part 2: Application to m2v2, a lower-body humanoid,'' \emph{The
  International Journal of Robotics Research}, vol.~31, no.~10, pp. 1117--1133,
  2012.

\bibitem{Posa2016}
M.~Posa, S.~Kuindersma, and R.~Tedrake, ``Optimization and stabilization of
  trajectories for constrained dynamical systems,'' in \emph{2016 IEEE
  International Conference on Robotics and Automation (ICRA)}, May 2016, pp.
  1366--1373.

\bibitem{Erez2013}
T.~Erez, K.~Lowrey, Y.~Tassa, V.~Kumar, S.~Kolev, and E.~Todorov, ``An
  integrated system for real-time model predictive control of humanoid
  robots,'' in \emph{2013 13th IEEE-RAS International Conference on Humanoid
  Robots (Humanoids)}, Oct 2013, pp. 292--299.

\bibitem{Koenemann2015}
J.~Koenemann, A.~D. Prete, Y.~Tassa, E.~Todorov, O.~Stasse, M.~Bennewitz, and
  N.~Mansard, ``Whole-body model-predictive control applied to the hrp-2
  humanoid,'' in \emph{2015 IEEE/RSJ International Conference on Intelligent
  Robots and Systems (IROS)}, Sept 2015, pp. 3346--3351.

\bibitem{Westervelt2007}
E.~R. Westervelt, C.~Chevallereau, J.~H. Choi, B.~Morris, and J.~W. Grizzle,
  \emph{Feedback Control of Dynamic Bipedal Robot Locomotion}.\hskip 1em plus
  0.5em minus 0.4em\relax Boca Raton, FL: CRC Press, 2007.

\bibitem{Hereid2018}
A.~Hereid, C.~M. Hubicki, E.~A. Cousineau, and A.~D. Ames, ``Dynamic humanoid
  locomotion: A scalable formulation for hzd gait optimization,'' \emph{IEEE
  Transactions on Robotics}, vol.~34, no.~2, pp. 370--387, April 2018.

\bibitem{Chevallereau2003}
C.~Chevallereau, G.~Abba, Y.~Aoustin, F.~Plestan, E.~R. Westervelt,
  C.~Canudas-De-Wit, and J.~W. Grizzle, ``Rabbit: a testbed for advanced
  control theory,'' \emph{IEEE Control Systems Magazine}, vol.~23, no.~5, pp.
  57--79, Oct 2003.

\bibitem{Chevallereau2009}
C.~Chevallereau, J.~W. Grizzle, and C.~Shih, ``Asymptotically stable walking of
  a five-link underactuated 3-d bipedal robot,'' \emph{IEEE Transactions on
  Robotics}, vol.~25, no.~1, pp. 37--50, Feb 2009.

\bibitem{Sreenath2011}
K.~Sreenath, H.-W. Park, I.~Poulakakis, and J.~W. Grizzle, ``A compliant hybrid
  zero dynamics controller for stable, efficient and fast bipedal walking on
  mabel,'' \emph{The International Journal of Robotics Research}, vol.~30,
  no.~9, pp. 1170--1193, 2011.

\bibitem{Hereid2014}
A.~Hereid, S.~Kolathaya, M.~S. Jones, J.~Van~Why, J.~W. Hurst, and A.~D. Ames,
  ``Dynamic multi-domain bipedal walking with atrias through slip based
  human-inspired control,'' in \emph{Proceedings of the 17th International
  Conference on Hybrid Systems: Computation and Control}, ser. HSCC '14.\hskip
  1em plus 0.5em minus 0.4em\relax New York, NY, USA: ACM, 2014, pp. 263--272.

\bibitem{Benbrahim1997}
H.~Benbrahim and J.~A. Franklin, ``Biped dynamic walking using reinforcement
  learning,'' \emph{Robotics and Autonomous Systems}, vol.~22, no.~3, pp. 283
  -- 302, 1997, robot Learning: The New Wave.

\bibitem{Lillicrap2015}
T.~P. Lillicrap, J.~J. Hunt, A.~Pritzel, N.~Heess, T.~Erez, Y.~Tassa,
  D.~Silver, and D.~Wierstra, ``Continuous control with deep reinforcement
  learning,'' \emph{CoRR}, vol. abs/1509.02971, 2015.

\bibitem{Schulman2015}
J.~Schulman, P.~Moritz, S.~Levine, M.~I. Jordan, and P.~Abbeel,
  ``High-dimensional continuous control using generalized advantage
  estimation,'' \emph{CoRR}, vol. abs/1506.02438, 2015.

\bibitem{Da2016}
X.~Da, O.~Harib, R.~Hartley, B.~Griffin, and J.~W. Grizzle, ``From 2d design of
  underactuated bipedal gaits to 3d implementation: Walking with speed
  tracking,'' \emph{IEEE Access}, vol.~4, pp. 3469--3478, 2016.

\bibitem{Da2017}
X.~{Da} and J.~{Grizzle}, ``{Combining Trajectory Optimization, Supervised
  Machine Learning, and Model Structure for Mitigating the Curse of
  Dimensionality in the Control of Bipedal Robots},'' \emph{ArXiv e-prints},
  Nov. 2017.

\bibitem{Todorov2012}
E.~Todorov, T.~Erez, and Y.~Tassa, ``Mujoco: A physics engine for model-based
  control,'' in \emph{2012 IEEE/RSJ International Conference on Intelligent
  Robots and Systems}, Oct 2012, pp. 5026--5033.

\bibitem{Ames2013Human}
A.~D. Ames, ``Human-inspired control of bipedal robots via control lyapunov
  functions and quadratic programs,'' in \emph{Proceedings of the
  16\textsuperscript{th} international conference on Hybrid systems:
  computation and control}, C.~Belta and F.~Ivancic, Eds., ACM.\hskip 1em plus
  0.5em minus 0.4em\relax {ACM}, 2013, pp. 31--32.

\bibitem{Mnih2013}
V.~{Mnih}, K.~{Kavukcuoglu}, D.~{Silver}, A.~{Graves}, I.~{Antonoglou},
  D.~{Wierstra}, and M.~{Riedmiller}, ``{Playing Atari with Deep Reinforcement
  Learning},'' \emph{ArXiv e-prints}, Dec. 2013.

\bibitem{Schulman2017}
J.~{Schulman}, F.~{Wolski}, P.~{Dhariwal}, A.~{Radford}, and O.~{Klimov},
  ``{Proximal Policy Optimization Algorithms},'' \emph{ArXiv e-prints}, July
  2017.

\bibitem{Gomez2005}
F.~Gomez and J.~Schmidhuber, ``Evolving modular fast-weight networks for
  control,'' in \emph{Artificial Neural Networks: Formal Models and Their
  Applications -- ICANN 2005}, W.~Duch, J.~Kacprzyk, E.~Oja, and
  S.~Zadro{\.{z}}ny, Eds.\hskip 1em plus 0.5em minus 0.4em\relax Berlin,
  Heidelberg: Springer Berlin Heidelberg, 2005, pp. 383--389.

\bibitem{schulman2015high}
J.~Schulman, P.~Moritz, S.~Levine, M.~Jordan, and P.~Abbeel, ``High-dimensional
  continuous control using generalized advantage estimation,'' \emph{arXiv
  preprint arXiv:1506.02438}, 2015.

\bibitem{Levine2013}
S.~Levine and V.~Koltun, ``Guided policy search,'' in \emph{Proceedings of the
  30th International Conference on Machine Learning}, ser. Proceedings of
  Machine Learning Research, S.~Dasgupta and D.~McAllester, Eds., vol.~28,
  no.~3.\hskip 1em plus 0.5em minus 0.4em\relax Atlanta, Georgia, USA: PMLR,
  17--19 Jun 2013, pp. 1--9.

\bibitem{Peng2017}
X.~B. Peng, G.~Berseth, K.~Yin, and M.~Van De~Panne, ``Deeploco: Dynamic
  locomotion skills using hierarchical deep reinforcement learning,'' \emph{ACM
  Trans. Graph.}, vol.~36, no.~4, pp. 41:1--41:13, July 2017.

\bibitem{Xie2018}
Z.~{Xie}, G.~{Berseth}, P.~{Clary}, J.~{Hurst}, and M.~{van de Panne},
  ``{Feedback Control For Cassie With Deep Reinforcement Learning},''
  \emph{ArXiv e-prints}, Mar. 2018.

\bibitem{Ortega1988}
R.~Ortega and M.~W. Spong, ``Adaptive motion control of rigid robots: a
  tutorial,'' in \emph{Proceedings of the 27th IEEE Conference on Decision and
  Control}, Dec 1988, pp. 1575--1584 vol.2.

\bibitem{Slotine1989}
J.-J.~E. Slotine and W.~Li, ``Composite adaptive control of robot
  manipulators,'' \emph{Automatica}, vol.~25, no.~4, pp. 509 -- 519, 1989.

\bibitem{Wang2007}
X.~song Wang, Y.~hu~Cheng, and W.~Sun, ``A proposal of adaptive pid controller
  based on reinforcement learning,'' \emph{Journal of China University of
  Mining and Technology}, vol.~17, no.~1, pp. 40 -- 44, 2007.

\bibitem{Manoonpong2007}
P.~Manoonpong, T.~Geng, T.~Kulvicius, B.~Porr, and F.~Wörgötter, ``Adaptive,
  fast walking in a biped robot under neuronal control and learning,''
  \emph{PLOS Computational Biology}, vol.~3, no.~7, pp. 1--16, 07 2007.

\bibitem{Brockman2016}
G.~{Brockman}, V.~{Cheung}, L.~{Pettersson}, J.~{Schneider}, J.~{Schulman},
  J.~{Tang}, and W.~{Zaremba}, ``{OpenAI Gym},'' \emph{ArXiv e-prints}, June
  2016.

\bibitem{video_link}
``{RABBIT simulation results in MoJoCo},''
  \url{https://www.youtube.com/watch?v=dhHMfnl7YlU}, accessed: 2018-10-03.

\bibitem{Plestan2003}
F.~Plestan, J.~W. Grizzle, E.~R. Westervelt, and G.~Abba, ``Stable walking of a
  7-dof biped robot,'' \emph{IEEE Transactions on Robotics and Automation},
  vol.~19, no.~4, pp. 653--668, Aug 2003.

\bibitem{Hereid2017}
A.~Hereid and A.~D. Ames, ``Frost∗: Fast robot optimization and simulation
  toolkit,'' in \emph{2017 IEEE/RSJ International Conference on Intelligent
  Robots and Systems (IROS)}, Sept 2017, pp. 719--726.

\bibitem{Pinto2017}
L.~{Pinto}, J.~{Davidson}, R.~{Sukthankar}, and A.~{Gupta}, ``{Robust
  Adversarial Reinforcement Learning},'' \emph{ArXiv e-prints}, Mar. 2017.

\end{thebibliography}
\end{document}